\documentclass[runningheads]{llncs}

\usepackage[T1]{fontenc}
\usepackage{graphicx}
\usepackage{amsmath}
\usepackage{amssymb}
\usepackage{url}
\usepackage{booktabs}

\usepackage[hidelinks]{hyperref}

\begin{document}

\title{Separating Diagnosis from Control: Auditable Policy Adaptation in Agent-Based Simulations with LLM-Based Diagnostics}

\titlerunning{Separating Diagnosis from Control}

\author{Shaoxin Zhong \and
Yuchen Su \and
Michael Witbrock}

\authorrunning{S. Zhong et al.}

\institute{University of Auckland, Auckland, New Zealand\\
\email{scho963@aucklanduni.ac.nz}}

\maketitle

\begin{abstract}
Mitigating elderly loneliness requires policy interventions that achieve both adaptability and auditability. Existing methods struggle to reconcile these objectives: traditional agent-based models suffer from static rigidity, while direct large language model (LLM) controllers lack essential traceability. This work proposes a three-layer framework that separates diagnosis from control to achieve both properties simultaneously. LLMs operate strictly as diagnostic instruments that assess population state and generate structured risk evaluations, while deterministic formulas with explicit bounds translate these assessments into traceable parameter updates. This separation ensures that every policy decision can be attributed to inspectable rules while maintaining adaptive response to emergent needs. We validate the framework through systematic ablation across five experimental conditions in elderly care simulation. Results demonstrate that explicit control rules outperform end-to-end black-box LLM approaches by 11.7\% while preserving full auditability, confirming that transparency need not compromise adaptive performance.
\keywords{Agent-based modeling \and Large language models \and Policy adaptation \and Auditable AI \and Social simulation}
\end{abstract}


\section{Introduction}

The mitigation of elderly loneliness constitutes a primary public health objective in aging societies, given its association with increased mortality risk and cognitive decline~\cite{who2025,oecd2025}. Effective policy interventions must be simultaneously \textit{adaptive}, responding to evolving population needs, and \textit{auditable}, offering transparent decision pathways for stakeholders.

Current approaches face a persistent trade-off between adaptivity and auditability. In policy domains such as elderly care, auditability is not merely desirable but often mandatory: stakeholders must understand why decisions were made to ensure accountability and maintain trust~\cite{rudin2019stop}. While Agent-Based Models (ABMs) effectively capture social dynamics and facilitate theory-driven simulation, policy interventions within them are often operationalized as pre-specified scenario rules or parameterized levers evaluated through comparative simulation rather than continuously re-optimized online~\cite{nespeca2023methodology,hammond2015abm}. Conversely, reinforcement learning enables policy adaptation through feedback-driven optimization, but many high-performing RL implementations rely on complex function approximators whose decision logic is difficult to directly inspect, creating an auditability gap in high-stakes policy settings~\cite{milani2024xrlsurvey}.

Large Language Models (LLMs) present an opportunity to bridge this gap through their capacity for reasoning and structured assessment. However, integrating LLMs into policy adaptation requires determining their optimal role. Should they function as direct controllers or as diagnostic support within a control framework? Existing LLM-augmented simulations predominantly adopt end-to-end architectures where LLMs act as direct decision-makers~\cite{park2023generative,gao2024large}, a design that retains the opacity of black-box systems. Although recent closed-loop frameworks have investigated structured feedback~\cite{shi2026coordinated,tan2024promptable}, the specific requirement for auditability in policy contexts remains under-addressed.

This paper introduces a framework that separates diagnosis from control by positioning LLMs as diagnostic instruments rather than direct policy actuators. Implemented via a three-layer architecture (Figure~\ref{fig:architecture}), the system ensures every decision remains traceable to explicit, inspectable rules. We validate this design through simulation of an elderly care facility with 30 agents over 200 days, comparing five experimental conditions including fixed policies and black-box LLM baselines.

Overall, our key contributions are as follows:
\begin{itemize}
    \item \textbf{Architecture.} We present a three-layer framework that achieves adaptive policy adjustment while maintaining full auditability by separating LLM diagnosis from deterministic control.
    \item \textbf{Design principles.} We validate three principles through systematic ablation: separation of diagnosis and control, explicit bounded update rules, and integration of theory-grounded simulation with data-driven diagnosis.
    \item \textbf{Empirical validation.} Results on holdout seeds demonstrate 15.3\% improvement over baseline and 11.7\% over black-box LLM control, showing that explicit formulas offer superior stability without compromising adaptability.
\end{itemize}

The remainder of this paper reviews related work (Section~\ref{sec:related}), details our methodology (Section~\ref{sec:method}), presents experimental results (Section~\ref{sec:experiments}), and discusses implications (Section~\ref{sec:discussion}) and limitations (Section~\ref{sec:limitations}).
\section{Related Work}
\label{sec:related}

\subsection{Agent-Based Models and LLM-Augmented Simulation}

Agent-based models effectively capture social dynamics and facilitate theory-driven simulation~\cite{grimm2020odd}. In elderly care contexts, ABMs model loneliness dynamics, social network evolution, and intervention effects. The core advantage is interpretability: agent rules explicitly encode modeller assumptions, enabling stakeholders to understand decision rationales. Traditional ABM interventions, however, are static. Parameters remain fixed throughout simulation runs.

Recent work has integrated LLMs into agent-based simulations through two approaches. The first uses LLMs to generate agent behaviors directly. Generative agents~\cite{park2023generative}, social interaction models~\cite{holt2025gsim}, and evolutionary simulations~\cite{jiang2024casevo} demonstrate that LLM-backed agents can exhibit emergent social dynamics. The second approach uses LLMs for higher-level reasoning. Land-system models~\cite{zeng2024exploring} and decision-making frameworks~\cite{kleiman2025simulation} position LLMs as institutional agents that interpret complex scenarios and generate strategic responses. Comprehensive surveys~\cite{gao2024large} map this rapidly expanding intersection, identifying persistent challenges in evaluation methodology and behavioral fidelity.

Critical work questions whether LLM agents reliably replicate human patterns~\cite{larooij2026critical}. The concern is twofold. First, generative flexibility does not guarantee faithful simulation of social dynamics. Second, evaluation frameworks remain underdeveloped, making it difficult to validate LLM agent behavior against empirical data. Alternative approaches using structured knowledge representations~\cite{muller2025kemass} guarantee auditability through explicit schemas but sacrifice the contextual flexibility needed to interpret emergent social signals.

Closed-loop architectures feed simulation state back into LLM decision processes. In pandemic control~\cite{shi2026coordinated}, traffic management~\cite{tan2024promptable}, and optimal control~\cite{fonseca2025optimal}, iterative refinement enables adaptive policy. These frameworks demonstrate technical viability. But they do not address auditability requirements central to social policy contexts. When LLMs directly control parameters, decisions cannot be traced to explicit rules.

\subsection{Positioning This Work}

The persistent challenge is reconciling adaptability with transparency. LLM-driven control adapts to emergent states but produces opaque decisions. Control-theoretic methods provide transparent optimization but lack semantic reasoning over complex agent states. No prior work separates diagnosis from control in this way. 
Our framework separates LLM-based diagnosis from deterministic control: 
LLMs serve strictly as diagnostic instruments, while explicit, 
bounded formulas translate their assessments into parameter updates.

\section{Methodology}
\label{sec:method}

\begin{figure}[t]
\centering
\includegraphics[width=1\textwidth]{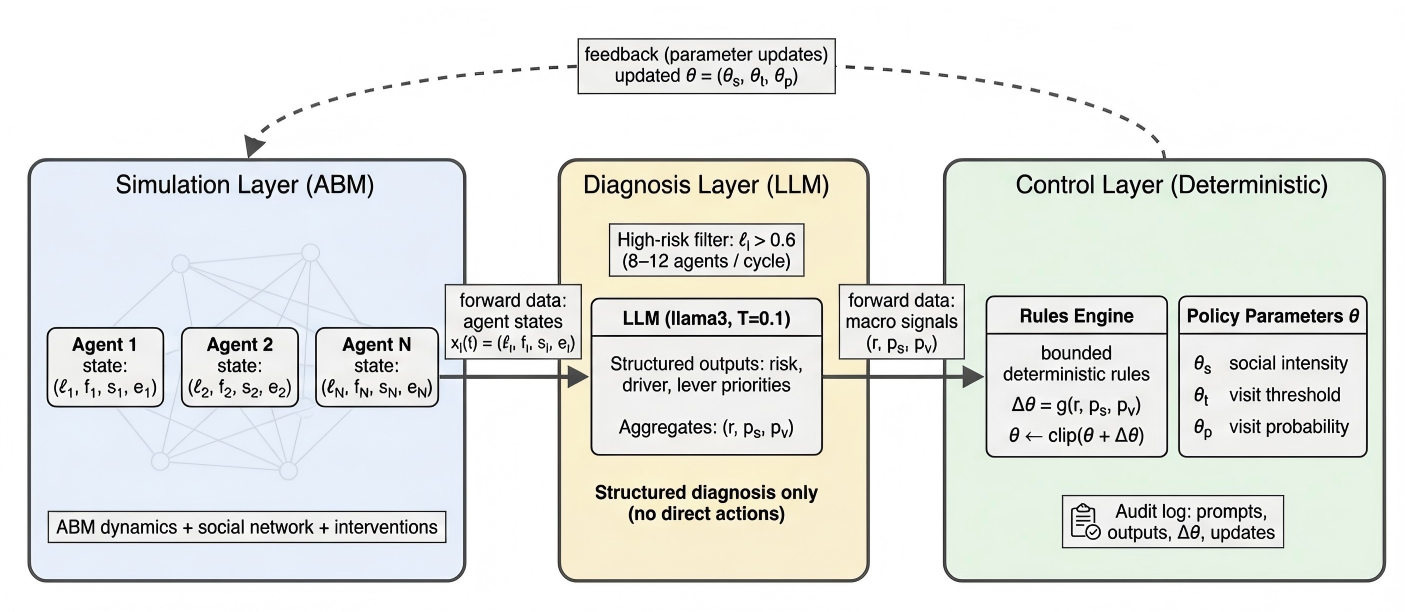}
\caption{Three-layer architecture for auditable policy adaptation. The simulation layer models agent dynamics, the diagnosis layer uses LLM-based diagnostics for risk assessment, and the control layer applies deterministic formulas. Solid arrows indicate data flow; the dashed arrow indicates the policy feedback loop.}
\label{fig:architecture}
\end{figure}

The framework addresses a core challenge in social policy adaptation: maintaining auditability while enabling responsive intervention adjustment. We achieve this through a three-layer architecture (Figure~\ref{fig:architecture}) that strictly separates semantic diagnosis (where LLMs operate) from policy control (where transparency is enforced). The approach is validated through simulation of an elderly care facility with $N=30$ agents over $T=200$ days, where interventions must adapt to evolving loneliness patterns while remaining traceable to explicit decision rules.

\subsection{Simulation Environment and Problem Formulation}

We model an elderly care facility where each agent $i$ is characterized by loneliness $\ell_i(t) \in [0,1]$, frailty $f_i(t)$, stress $s_i(t)$, and energy $e_i(t)$. Agents are embedded in a dynamic social network $G(t)$ where edges represent social ties. Network evolution follows homophily principles: agents with similar loneliness levels are more likely to form connections. This design reflects established social science theories while enabling controlled experimentation.

Two intervention types are available. Social events controlled by intensity $\theta_s \in [0.8, 1.5]$ reduce loneliness for all participants proportionally. Home visits provide targeted support for eligible agents (those with $\ell_i > \theta_t$), reducing both loneliness and stress with probability $\theta_p \in [0.15, 0.5]$. The eligibility threshold $\theta_t \in [0.4, 0.6]$ determines which agents receive visits. The objective is to minimize population-level loneliness (mean across all agents at final timestep) while maintaining full traceability of all policy decisions.

\subsection{Three-Layer Architecture for Auditable Adaptation}

The framework partitions responsibilities across three layers, each with a distinct role in the adaptation process. This separation is motivated by the requirement that policy decisions must be both responsive to emergent population needs and traceable to explicit rules that stakeholders can verify.

\textbf{Simulation Layer: Theory-Grounded Agent Dynamics.}
An agent-based model implements dynamics grounded in social science theories. Loneliness evolves toward individual baseline with rate $\alpha_\ell$ and decreases with social interactions at rate $\beta_\ell$. Frailty accumulates gradually with age and stress. Network formation follows homophily principles: potential ties form based on similarity in loneliness levels and existing network degree. Social events reduce loneliness for all participants, with effectiveness proportional to $\theta_s$. Home visits provide targeted support for eligible agents, reducing both loneliness and stress with probability $\theta_p$. This layer operates deterministically given fixed parameters, ensuring that variance across experimental runs stems only from network initialization.

\textbf{Diagnosis Layer: LLM-Based Risk Assessment.}
Every 7 simulation days, an LLM assesses high-risk agents to inform control decisions. To manage computational costs, only agents with loneliness exceeding 0.6 are diagnosed (typically 8 to 12 agents per cycle). For each high-risk agent, a structured prompt is constructed including current state, recent interaction history, and network position. The LLM (Ollama llama3:8b, temperature 0.1) produces structured JSON outputs containing risk level, primary driver identification, and intervention priorities for each policy lever. The use of local LLM deployment ensures reproducibility and avoids external API dependencies.

Individual assessments are aggregated to population-level statistics. The high-risk proportion $r$ represents the fraction of agents with assessed risk exceeding 0.6. Mean priority scores for social events ($p_s$) and home visits ($p_v$) are computed by averaging over diagnosed agents $H$:
\begin{equation}
p_s = \frac{1}{|H|} \sum_{i \in H} \text{priority}_i^{\text{social}}, \quad
p_v = \frac{1}{|H|} \sum_{i \in H} \text{priority}_i^{\text{visits}} 
\label{eq:priority}
\end{equation}
This aggregation provides a privacy-preserving interface to the control layer. Individual diagnoses are not directly visible to control formulas, ensuring that policy decisions operate on population-level signals rather than individual-level data.

\textbf{Control Layer: Deterministic Parameter Updates.}
Deterministic formulas convert population-level statistics into policy parameter updates. These update rules follow the logic of stepped and adaptive care: intervention intensity is adjusted using explicit decision rules at predefined stages based on monitored risk indicators and their cutoffs~\cite{almirall2014smart,wilhelm2024decisionrules}. Such cutoff-based rules are widely used to support reproducible, auditable escalation while balancing resource constraints and the risks of over- or under-treatment~\cite{wilhelm2024decisionrules}. In our simulation, we instantiate this principle with the thresholds $r>0.4$ (substantial high-risk prevalence) and $p_s>0.75$ (high priority on a normalized $[0,1]$ scale), which we treat as modeling choices rather than clinical constants. These values were selected through pilot experiments on train seeds to balance responsiveness and stability; their robustness is further validated via sensitivity analysis in Appendix~\ref{app:sensitivity}.
Update magnitudes are deliberately small, reflecting incremental adjustment:
\begin{align}
\Delta\theta_s &= \begin{cases}
\min(0.05,\; 0.1 \cdot p_s) & \text{if } r > 0.40 \text{ and } p_s > 0.75\\
0 & \text{otherwise}
\end{cases} \label{eq:social}\\
\Delta\theta_t &= \begin{cases}
-0.02 & \text{if } p_v > 0.75 \text{ and } \theta_t > 0.4\\
0 & \text{otherwise}
\end{cases} \nonumber\\
\Delta\theta_p &= \begin{cases}
+0.05 & \text{if } p_v > 0.75 \text{ and } \theta_p < 0.5\\
0 & \text{otherwise}
\end{cases} \label{eq:visit}
\end{align}

To promote stability, we bound each parameter update ($\lVert\Delta\theta\rVert 
\le 0.05$) as a conservative step-size control mechanism, implementing gradual 
escalation instead of abrupt changes. This is consistent with adaptive-intervention 
practice that emphasizes staged, monitored adjustments~\cite{almirall2014smart} 
and with evidence that changes in stimuli and daily activity patterns in 
nursing-home environments can affect resident outcomes, motivating cautious, 
incremental modifications~\cite{knippenberg2022stimuli}. After computing updates, 
parameters are clipped to remain within valid ranges. The separation of diagnosis 
(where neural computation occurs) from control (where deterministic rules operate) 
ensures that every parameter change can be attributed to specific macro statistics 
and verified against explicit criteria.

\section{Simulation Experiments}
\label{sec:experiments}

\subsection{Implementation and Experimental Setup}

To evaluate the proposed three-layer framework, we developed a simulation environment based on the theoretical dynamics described in Section~\ref{sec:method}.

We implemented the system in Python 3.8+ using NumPy for numerical computation and NetworkX for graph operations. Agent states update in discrete daily timesteps. Network edges are stored as adjacency lists and updated incrementally when homophily-based attachment rules trigger new tie formation. We run the core simulation logic on the CPU, while offloading LLM inference tasks to the GPU for efficiency.

For diagnosis, the system uses Ollama with \texttt{llama3:8b} for local inference. We construct structured prompts programmatically using JSON templates that inject agent state variables, interaction counts from the past 7 days, and network degree statistics. Responses are parsed using Python's \texttt{json} module and validated against a predefined schema. We set the temperature to 0.1 to minimize variance while preserving structured reasoning. Each diagnosis call takes approximately 0.2 seconds (median).

The control layer functions as a deterministic controller. It accepts aggregated statistics ($r$, $p_s$, $p_v$) and current parameters ($\theta_s$, $\theta_t$, $\theta_p$), then returns updates according to Equations~\ref{eq:social}--\ref{eq:visit}. This logic uses explicit conditional branches with no learned components, ensuring reproducibility and enabling formal verification.

We initialized the environment with a population of $N=30$ agents over $T=200$ days. Each run takes approximately 15 to 20 minutes, dominated by simulation stepping and diagnosis cycles. To ensure consistent network structures, we used fixed random seeds for all experiments. We used seeds 42, 100, and 200 for development, while the reported results are based on four holdout seeds (300, 400, 500, 600) to validate performance on unseen initializations.
\begin{table}[t]
\caption{Experimental conditions.}
\label{tab:conditions}
\centering
\small
\begin{tabular}{@{}lccc@{}}
\toprule
\textbf{Condition} & \textbf{LLM Diagnosis} & \textbf{Adaptation} & \textbf{Control Type} \\
\midrule
Baseline & -- & -- & None \\
Fixed Policy & -- & -- & Static $\theta$ \\
LLM Mapping & \checkmark & -- & Fixed response \\
Closed-loop (Ours) & \checkmark & \checkmark & Deterministic \\
Black-box LLM & \checkmark & \checkmark & LLM decides $\theta$ \\
\bottomrule
\end{tabular}
\end{table}

Baseline runs with no intervention. Fixed Policy applies static parameters ($\theta_s=1.0$, $\theta_t=0.6$, $\theta_p=0.3$) throughout. LLM Mapping activates diagnosis but uses a fixed switching rule: social intensity is 1.2 if high-risk proportion exceeds 0.4, otherwise 1.0. This isolates semantic assessment without adaptive updates. Closed-loop implements the full framework (Equations~\ref{eq:social}--\ref{eq:visit}). Black-box LLM prompts the LLM to directly suggest parameter values given macro statistics ($r$, $p_s$, $p_v$). Parameter bounds are enforced, but decision logic is not specified by explicit rules.

Table~\ref{tab:results} shows holdout performance across all conditions. Closed-loop achieves the lowest mean loneliness (0.607), corresponding to a 15.3\% reduction over baseline (0.717). Fixed Policy reduces loneliness by 6.0\%, while LLM Mapping and Black-box LLM achieve 5.4\% and 4.2\% reductions respectively.

\begin{table}[t]
\caption{Holdout performance across four seeds.}
\label{tab:results}
\centering
\begin{tabular}{@{}lcc@{}}
\toprule
\textbf{Method} & \textbf{Mean} & \textbf{SD} \\
\midrule
Baseline & 0.717 & 0.018 \\
Fixed Policy & 0.674 & 0.012 \\
LLM Mapping & 0.680 & 0.007 \\
\textbf{Closed-loop (Ours)} & \textbf{0.607} & \textbf{0.020} \\
Black-box LLM & 0.687 & 0.025 \\
\bottomrule
\end{tabular}
\end{table}

Pairwise comparisons (Table~\ref{tab:pairwise}) reveal several patterns. Closed-loop significantly outperforms all alternatives: 10.7\% better than LLM Mapping ($p<.001$), 11.7\% better than Black-box LLM ($p=.002$), and 10.0\% better than Fixed Policy ($p=.001$). LLM Mapping shows no improvement over Fixed Policy ($p=0.511$). Black-box LLM also shows no significant gain over Fixed Policy ($p=0.385$). Effect sizes are large ($d > 3.5$), reflecting the low within-condition variance (SD = 0.007 to 0.025) characteristic of deterministic simulation dynamics where variance stems primarily from network initialization across seeds.


\begin{table}[t]
\centering
\caption{Pairwise comparisons. Large effect sizes reflect deterministic simulation dynamics.}
\label{tab:pairwise}
\resizebox{0.95\columnwidth}{!}{

\begin{tabular}{@{}lcccc@{}}
\toprule
\textbf{Comparison} & \textbf{Mean Diff.($\Delta$)} & \textbf{Improve(\%)} & \textbf{Cohen's($d$)} & \textbf{$p$-value} \\
\midrule
Closed vs.\ LLM Mapping & $-0.073$ & $10.7\%$ & $-4.87$ & $< .001^{***}$ \\
Closed vs.\ Black-box   & $-0.081$ & $11.7\%$ & $-3.54$ & $ .002^{**}$ \\
Closed vs.\ Fixed       & $-0.067$ & $10.0\%$ & $-4.05$ & $ .001^{**}$ \\
Black-box vs.\ Fixed    & $+0.013$ & --       & $+0.66$ & $ .385$ \\
\bottomrule
\multicolumn{5}{l}{\small $^{***}p<.001$, $^{**}p<.01$}
\end{tabular}
}

\end{table}

Pairwise analysis (Table~\ref{tab:pairwise}) treats each seed as an independent unit of analysis ($n=4$ per condition). Effect sizes are computed using Cohen's $d$ with pooled standard deviations across holdout seeds. $p$-values are obtained via two-sample t-tests comparing seed-level final loneliness distributions between conditions. Large effect sizes ($d > 3.5$) reflect the deterministic nature of ABM simulations: given identical parameters, the simulation produces identical trajectories, yielding low within-condition variance (SD $= 0.007$ to $0.025$).

\begin{figure}[t]
\centering
\includegraphics[width=0.85\textwidth]{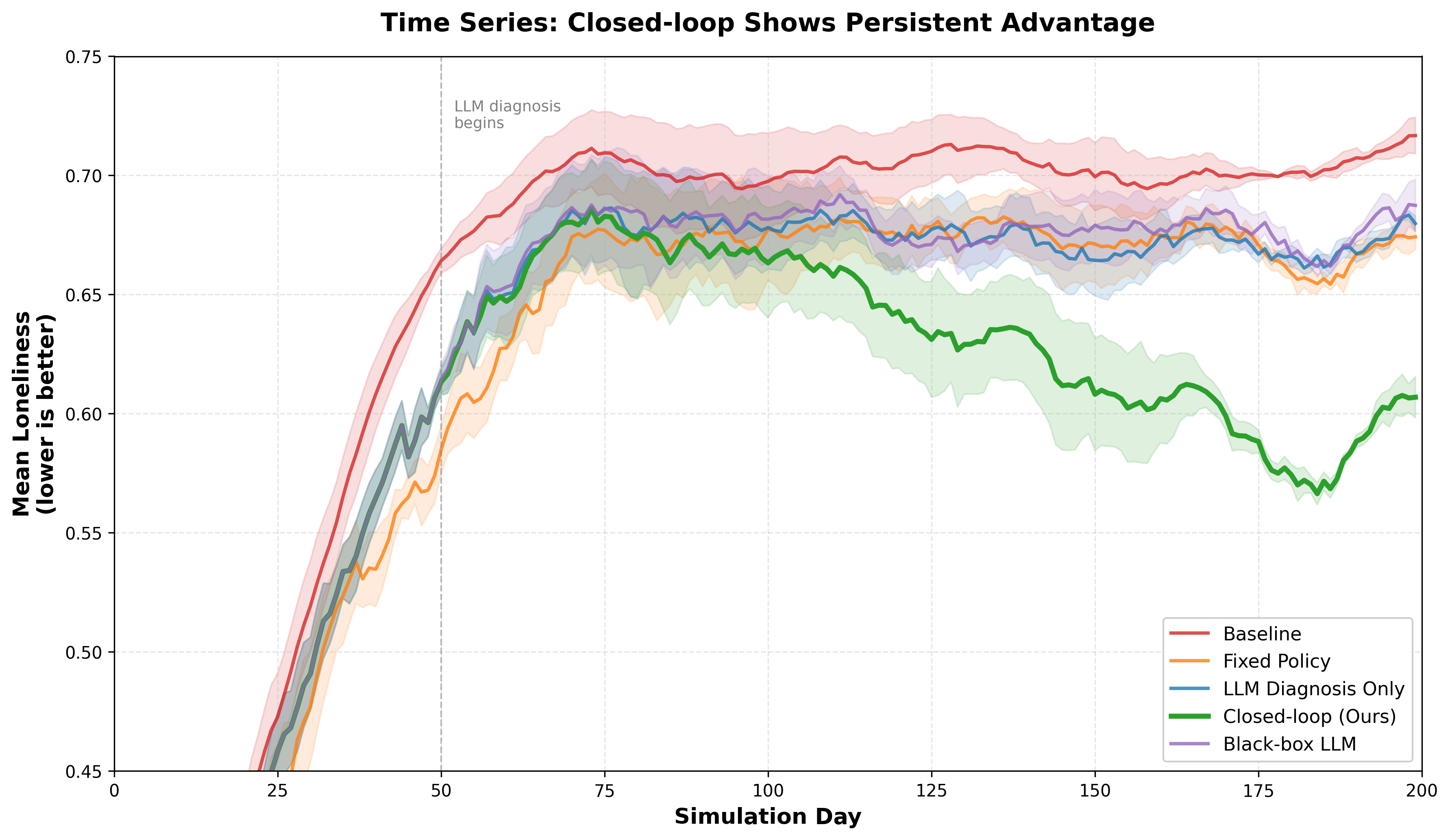}
\caption{Loneliness trajectories over 200 days ($N=30$). Shaded regions: $\pm 1$ SD across 4 seeds.}
\label{fig:timeseries}
\end{figure}

Loneliness trajectories over 200 days (Figure~\ref{fig:timeseries}) show how conditions diverge over time.

\begin{figure}[t!]
\centering
\includegraphics[width=\textwidth]{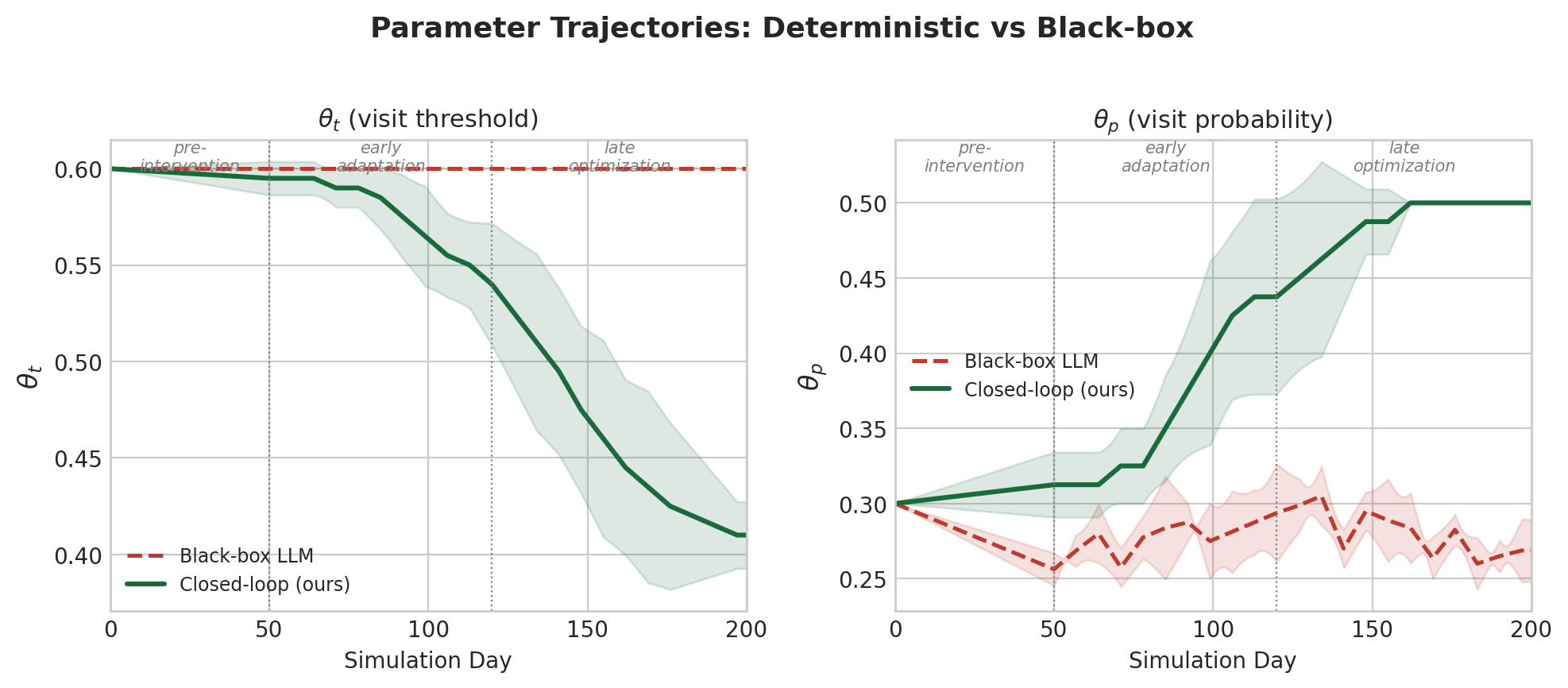}
\caption{Parameter trajectories. Shaded regions: $\pm 1$ SD across 4 seeds. (Left) Visit threshold $\theta_t$. (Right) Visit probability $\theta_p$.}
\label{fig:mechanisms}
\end{figure}

Parameter trajectories (Figure~\ref{fig:mechanisms}) reveal how the deterministic controller adapts. Social event intensity ($\theta_s$) remains at 1.0 throughout all runs; population-level social priority never exceeds the required threshold. In contrast, home visit parameters adapt systematically. Across all seeds, $\theta_t$ decreases from 0.6 to 0.4 and $\theta_p$ increases from 0.3 to 0.5. The controller reaches the upper bound for $\theta_p$ (0.50) by Day 134 and the lower bound for $\theta_t$ (0.40) by Day 176. Black-box LLM shows higher cross-seed variance and less systematic progression. Parameters oscillate rather than converge monotonically to bounds.

\section{Discussion}
\label{sec:discussion}

The closed-loop approach achieves 15.3\% loneliness reduction over baseline and outperforms all comparison conditions on holdout seeds. The most striking result is the 11.7\% gap between closed-loop and black-box LLM. This raises a question: why do explicit formulas outperform an LLM that can flexibly reason about population state?

Two mechanisms explain the gap. First, bounded updates prevent instability. The deterministic controller (Equations~\ref{eq:social}--\ref{eq:visit}) caps each parameter change at 0.05 per update cycle. When high-risk proportion $r$ exceeds 0.4 and social priority $p_s$ exceeds 0.75, social intensity increases by at most 0.05. The system then responds over 7 days before the next update. Black-box LLM faces no such constraint. At each cycle, it can suggest any value within bounds. This produces larger adjustments and occasional reversals. The consequence appears in outcome variance: SD = 0.025 for black-box versus 0.020 for closed-loop. Parameter trajectories (Figure~\ref{fig:mechanisms}) visualize the difference. Closed-loop parameters move monotonically toward bounds. Black-box oscillates.

Second, explicit rules encode domain constraints reliably. Equation~\ref{eq:social} implements a conjunction: increase intensity only if $r > 0.4$ AND $p_s > 0.75$. This is deterministic. Given identical inputs, the controller always produces identical outputs. Black-box LLM receives natural language prompts describing these thresholds. It must reconstruct the logic at every cycle. Small variations in how the LLM parses the prompt lead to inconsistent decisions for similar states. In policy contexts where decisions must be reproducible and auditable, this consistency matters.

The comparison between Black-box LLM (0.687) and Fixed Policy (0.674) sharpens the point. Black-box performs worse than doing nothing adaptive at all. Adaptation without structural constraints degrades performance through instability. This is not specific to our model or prompt. It reflects a broader issue: language models trained on text corpora are not optimized for discrete control tasks requiring stable, incremental adjustment.
\section{Limitations and Future Directions}
\label{sec:limitations}

The experiments use a small population ($N=30$) with simplified loneliness dynamics. Real facilities house 50 to 200 residents, and loneliness is affected by factors not modeled here: family contact frequency, acute health events, and staffing constraints. Direct validation against real facility data was not feasible due to privacy restrictions. Testing larger populations and calibrating dynamics against anonymized measurements would strengthen external validity.

We evaluate one local LLM configuration (Ollama llama3:8b) with a single prompt structure. Different models or prompting strategies may yield different diagnosis quality. That said, the core design choice (using LLMs for diagnosis while keeping control explicit) does not depend on any specific model. Evaluating alternative architectures like GPT-4 or Claude would test generalization.

The update rules in Equations~\ref{eq:social}--\ref{eq:visit} are hand-specified. Future work could learn rule parameters from data while preserving interpretability, for example via symbolic regression constrained to simple functions~\cite{fonseca2025optimal} or rule learning from historical intervention data. This would reduce manual specification while maintaining auditability.

We report intervention costs (LLM calls, visit counts) but do not optimize cost directly. Practical deployment requires balancing loneliness reduction against resource constraints. Extending the framework to multi-objective optimization is important for real-world adoption, particularly in settings where budget limitations are strict.

Although the case study focuses on elderly care, the design principles apply to other policy domains requiring both adaptation and transparency. Healthcare resource allocation, educational interventions, and urban planning all face similar trade-offs between flexible response and stakeholder accountability. Testing the architecture in these domains would validate generalizability and identify domain-specific challenges.

\section{Conclusion}
\label{sec:conclusion}

We presented an auditable closed-loop framework for policy adaptation in agent-based simulations. The key design choice is positioning LLMs as diagnostic tools and implementing parameter adaptation through explicit deterministic rules.

The ablation study yields three main findings. First, the closed-loop condition achieves 15.3\% reduction in final loneliness relative to baseline, outperforming all alternative conditions on holdout seeds. Second, deterministic control outperforms end-to-end LLM parameter control by 11.7\%. Explicit rules improve stability and decision consistency, showing that auditability can be achieved without sacrificing performance. Third, diagnosis alone offers little benefit when not coupled with adaptive control. LLM Mapping performs no better than fixed policies, indicating that diagnostic signals must be translated into structured parameter updates to produce operational gains.

Overall, the results support a practical principle for high-stakes applications: LLMs are better suited to structured assessment, while policy control should remain explicit and inspectable. This separation offers a viable path to deploy adaptive AI components in policy domains where accountability is non-negotiable.




\appendix
\section{Parameter Sensitivity Analysis}
\label{app:sensitivity}
\begin{table}[ht]
\centering
\caption{Sensitivity on holdout seeds 300, 400 (Baseline: $r=0.4$, $p=0.75$, cap$=0.05$).}
\label{tab:sensitivity-app}
\small
\begin{tabular*}{\linewidth}{@{\extracolsep{\fill}}lcccccc@{}}
\toprule
\textbf{Parameter} & \textbf{Value} & \textbf{Mean} & \textbf{SD} & \textbf{Min} & \textbf{Max} & \textbf{$\Delta$\%} \\
\midrule
\textbf{Baseline} & -- & \textbf{0.636} & 0.016 & 0.623 & 0.648 & -- \\
\midrule
Risk threshold ($r$) & 0.30 & 0.636 & 0.016 & 0.623 & 0.648 & +0.0\% \\
                     & 0.50 & 0.636 & 0.016 & 0.623 & 0.648 & +0.0\% \\
\midrule
Priority ($p$)       & 0.65 & 0.636 & 0.016 & 0.623 & 0.648 & +0.0\% \\
                     & 0.85 & 0.680 & 0.007 & 0.677 & 0.682 & +6.4\% \\
\midrule
Update cap ($\Delta\theta$) & 0.03 & 0.636 & 0.016 & 0.623 & 0.648 & +0.0\% \\
                            & 0.08 & 0.636 & 0.016 & 0.623 & 0.648 & +0.0\% \\
\bottomrule
\end{tabular*}
\end{table}
Outcomes are insensitive to $r$ and cap in the tested ranges; only $p=0.85$ shifts results (+6.4\%), so $p=0.75$ is retained as default.

\section{LLM Diagnosis Examples}
\label{app:llm}
Ollama \texttt{llama3} ($T=0.1$) assesses agents every 7 days, outputting JSON risk labels and intervention priorities.

\begin{table}[ht]
\centering
\caption{Example diagnosis outputs (Day 120, Seed 300; 30 agents diagnosed).}
\label{tab:llm-example}
\small
\begin{tabular}{@{}clllc@{}}
\toprule
\textbf{Agent} & \textbf{Risk (L)} & \textbf{Risk (F)} & \textbf{Primary Driver} & \textbf{Priority} \\
\midrule
Agent 8  & High   & Low    & Social isolation    & Network: 0.80 \\
Agent 13 & High   & Medium & Health \& isolation & Network: 0.80 \\
Agent 0  & Medium & Low    & Minor mood dip      & Visit: 0.70 \\
\midrule
\multicolumn{5}{l}{\small \textit{7/30 classified as High risk ($r=0.23$)}} \\
\bottomrule
\end{tabular}
\end{table}
Diagnosis outputs feed into Eqs.~\ref{eq:social}--\ref{eq:visit}; 
here $r=0.23 < 0.4$ (Appendix~\ref{app:sensitivity}), so social intensity remains unchanged.

\end{document}